 \documentclass[preprint,12pt]{elsarticle}

\usepackage{amssymb}
\usepackage{amsmath}

\usepackage[colorlinks=true,linkcolor=blue,citecolor=blue,urlcolor=blue]{hyperref}
\usepackage{booktabs} 
\usepackage{multirow}
\usepackage{makecell}
\usepackage{subcaption}  
\usepackage{threeparttable} 
\usepackage{makecell}
\usepackage{array}
\usepackage{enumitem}
\usepackage{float}
\usepackage{longtable}
\usepackage{tabularx,adjustbox,pifont}
\newcommand{\cmark}{\ding{51}} 
\newcommand{\xmark}{\ding{55}} 
\newcolumntype{C}[1]{>{\centering\arraybackslash}m{#1}}
\newcolumntype{L}[1]{>{\raggedright\arraybackslash}m{#1}}
\newcolumntype{R}[1]{>{\raggedleft\arraybackslash}m{#1}} 
\usepackage{tabularx}
\usepackage{adjustbox}
\newcolumntype{Y}{>{\raggedright\arraybackslash}X}
\usepackage{pifont}
\usepackage{amsmath}
\usepackage{hyperref}

\journal{Artificial Intelligence in Medicine}
\usepackage{caption}
\begin{document}

\begin{frontmatter}



\title{XAttn-BMD: Multimodal Deep Learning with Cross-Attention for Femoral Neck Bone Mineral Density Estimation}


\author[ECS]{Yilin Zhang}
\author[HDH]{Leo D. Westbury}
\author[HDH]{Elaine~M.~Dennison}
\author[HDH,NIHR]{Nicholas~C.~Harvey}
\author[HDH]{Nicholas~R.~Fuggle}
\author[ECS]{Rahman~Attar\corref{cor1}}

\cortext[cor1]{Corresponding author: r.attar@southampton.ac.uk}
\address[ECS]{School of Electronics and Computer Science, University of Southampton, UK}
\address[HDH]{MRC Lifecourse Epidemiology Centre, University of Southampton, Southampton General Hospital, UK}
\address[NIHR]{NIHR Southampton Biomedical Research Centre, University of Southampton and University Hospital NHS Foundation Trust, Southampton, UK}

\begin{abstract}
Poor bone health is a significant public health concern, and low bone mineral density (BMD) leads to an increased fracture risk, a key feature of osteoporosis. We present XAttn-BMD (Cross-Attention BMD), a multimodal deep learning framework that predicts femoral neck BMD from hip X-ray images and structured clinical metadata. It utilizes a novel bidirectional cross-attention mechanism to dynamically integrate image and metadata features for cross-modal mutual reinforcement. A Weighted Smooth L1 loss is tailored to address BMD imbalance and prioritize clinically significant cases. Extensive experiments on the data from the Hertfordshire Cohort Study show that our model outperforms the baseline models in regression generalization and robustness. Ablation studies confirm the effectiveness of both cross-attention fusion and the customized loss function. Experimental results show that the integration of multimodal data via cross-attention outperforms naive feature concatenation without cross-attention, reducing MSE by 16.7\%, MAE by 6.03\%, and increasing the \(R^2\) score by 16.4\%, highlighting the effectiveness of the approach for femoral neck BMD estimation. Furthermore, screening performance was evaluated using binary classification at clinically relevant femoral neck BMD thresholds, demonstrating the model's potential in real-world scenarios.
\end{abstract}


\end{frontmatter}



\section{Introduction}
Bone mineral density (BMD) is a key determinant of bone health, with low BMD being a major risk factor for osteoporotic fractures and associated mortality in aging populations\cite{friedman2006important, nguyen2004osteoporosis, clynes2020epidemiology, fonseca2014bone, johnell2005predictive, bolland2010effect, tolar2004osteopetrosis}. Osteoporotic fractures impose substantial personal, societal, and economic burdens, and relationship status has been linked to adverse health outcomes and hip fracture risk\cite{liu2024multi}. It often asymptomatic until a fracture occurs\cite{ghalenavi2024diagnostic}, is diagnosed based on BMD measurements\cite{fonseca2014bone, johnell2005predictive}. Among measurement sites, femoral neck BMD is widely recognized as one of the most predictive indicators of osteoporotic fracture risk, due to its strong association with hip fractures and related complications\cite{dargent1996fall}. Dual-energy X-ray absorptiometry (DXA) is the clinical standard for assessing BMD\cite{dimai2017use, messina2016dual}. However, the high cost and limited accessibility of DXA in some developing countries restrict its widespread use\cite{ghalenavi2024diagnostic}. In such contexts, X-ray imaging, which is far more widely available, offers a practical alternative for BMD assessment, facilitating early diagnosis and screening in resource-limited settings.

Recent studies have explored deep learning approaches for BMD estimation from X-ray images. However, existing methods often have limitations in both performance and clinical applicability. A major limitation is the scarcity and imbalance of annotated medical datasets, where clinical cases related to diseases are underrepresented, making it difficult for models to generalize. In addition, for measuring femoral neck BMD, routinely acquired hip X-ray images vary in quality, often lacking contrast and anatomical consistency. Importantly, most existing methods rely exclusively on visual features extracted from medical images, overlooking the valuable contextual information embedded in clinical metadata, which is highly correlated with BMD measurements. Leveraging these variables enables clinical factor–level interpretability by quantifying each factor’s contribution to the estimation.

To address these challenges, we propose \textbf{XAttn-BMD} (Cross-Attention BMD), which is a lightweight multimodal regression framework based on cross-attention mechanisms that jointly utilizes visual and structured metadata features for reliable femoral neck BMD estimation. The framework uses ResNet34 as the backbone network to extract image features and encodes metadata features through a multilayer perceptron (MLP). The multi-layer, multi-head cross-attention module enables bidirectional interaction between the two modalities (image and metadata), thereby enhancing feature representation through mutual regulation. We then concatenate the cross-attended representations before the regression head. Experimental results show that compared with directly fusing image features and metadata features, this method achieves better prediction performance.

Furthermore, to address the imbalanced BMD distribution in clinical data, this study introduces a weighted Smooth L1 loss to place greater emphasis on high and low BMD values, which are often underrepresented yet clinically important. This improves the model's sensitivity to low bone mass (BMD 1.0–2.5 SD below the young-adult female reference) and enhances the potential for risk screening.

Building on these advances, this study contributes to both engineering and clinical domains. From an engineering perspective, it presents a lightweight and effective fusion architecture suitable for small-scale, imbalanced regression tasks, which is a common challenge in medical datasets. From a clinical perspective, the model demonstrates strong potential for predicting femoral neck BMD, supporting early identification of individuals at risk of low bone mass and osteoporosis, thereby enabling timely diagnosis and preventive care, particularly in resource-limited settings. To further evaluate the clinical utility of our approach, we assessed model performance in a binary classification task using established clinical thresholds for low BMD, providing direct insight into its potential role in osteoporosis risk screening.

Overall, this study focuses on estimating femoral neck BMD and screening for low bone mass. The main contributions of this work are summarized as follows:

1. A multimodal cross-attention framework enabling bidirectional cross-modal interaction between image and metadata features, yielding richer representations and clinical factor-level interpretability for femoral neck BMD estimation.

2. A Weighted Smooth L1 loss to address data imbalance and emphasize clinically important low femoral neck BMD values, improving the model's robustness to underrepresented but medically critical cases.

3. Extensive experiments demonstrate that the proposed method achieves superior performance compared to conventional concatenation across multiple evaluation metrics.

4. Clinical screening performance evaluation through binary classification at clinically meaningful femoral neck BMD thresholds, demonstrating the model's potential for real-world low bone mass risk stratification.

The code of this study is available at: \url{https://github.com/YilinZhang00/BMD-regression}

\section{Related Work}
Modern imaging technology combined with machine learning algorithms provides new perspectives for early diagnosis and risk assessment of osteoporosis, with broad clinical application prospects. Machine learning approaches for BMD estimation have evolved significantly, with early studies demonstrating the potential of computational methods for osteoporosis risk assessment \cite{lu2001classification, kim2013osteoporosis}. For instance, Lu et al. showed that models based on elderly populations can significantly improve osteoporosis diagnosis consistency, while combining BMD assessments from multiple regions effectively enhances hip and spine fracture risk prediction \cite{lu2001classification}. Subsequently, Kim et al. demonstrated that support vector machines achieved superior performance compared to traditional tools for identifying high-risk populations \cite{kim2013osteoporosis}.

Recent advances in deep learning have opened new possibilities for automated BMD estimation from medical images. In 2023, Feng et al. combined U-Net for image segmentation with DenseNet121 for classification, achieving 74\% classification accuracy \cite{feng2023deep}. And Zhang et al. proposed a joint learning framework that combines localization, segmentation, and classification. A test classification accuracy of 93.3\% demonstrates the potential of this method in the diagnosis of osteoporosis \cite{zhang2023end}. There are also studies on multiple sites of osteoporosis. Another study in 2023 used CNN to estimate the bone density of the femoral neck from standard hand, knee, and pelvic X-rays. The results showed that the model can effectively screen patients with low bone density even if the femoral neck is not in the field of view. Experimental results also indicated that the sensitivity and specificity of the model for low bone density classification are both over 75\%, and X-ray images and the co-variate data contribute equally to the model performance \cite{golestan2023approximating}. Additional studies have demonstrated that deep learning models trained on large-scale X-ray collections can provide accurate BMD estimates, further supporting the clinical utility of AI-based approaches for osteoporosis screening \cite{ho2021application, hsieh2021automated}.

Current studies have explored the use of deep learning to predict BMD from traditional radiological images, providing an alternative to DXA in resource-limited regions. For example, a two-tier stacked regression method was used in a radiological model based on abdominal computed tomography (CT). The results showed that its predictive performance was better than that of a single regression model and was highly correlated with the results of dual-energy X-ray absorptiometry (DXA), suggesting CT images are of great value in the diagnosis of osteoporosis \cite{dai2023radiomics}. Furthermore, Sato et al. conducted a multicenter study to estimate lumbar spine BMD and T-score values using chest X-rays \cite{sato2022deep}. Similarly, Wang et al. \cite{wang2022lumbar} proposed a multi-ROI modeling strategy combined with an anatomically aware attention mechanism to improve the accuracy of BMD estimation from chest X-rays. In addition, Galbusera et al. \cite{galbusera2024estimating} explored BMD estimation using traditional spinal MRI and X-rays. While these methods have demonstrated the feasibility of non-DXA BMD  estimation, they mainly rely on image features without leveraging informative clinical or metadata inputs.

Beyond single-modality approaches, multimodal fusion strategies have emerged to combine different modalities to enhance regression performance. Cross-attention is a variant of the attention architecture where the query comes from one modality and the key/value comes from another modality and has been shown to effectively capture interactions across heterogeneous data sources \cite{saioni-giannone-2024-multimodal}. Benefiting from the cross-attention mechanism, Yang et al. proposed a graph-based adaptive fusion framework combined with phenotypic data to leverage multimodal features for Alzheimer's disease diagnosis \cite{yang2023multi}. Similarly, Zhang et al. \cite{zhang2023multi}, focusing on the same disease, developed a cross-attention-based architecture for predicting Alzheimer's that combined structural magnetic resonance imaging (sMRI), fluorodeoxyglucose-positron emission tomography (FDG-PET) and cerebrospinal fluid (CSF) biomarkers. Additionally, Liu et al. \cite{liu2024multi} utilized genomic data and pathological images, proposing a fusion network with both intra-modality and inter-modality attention for breast cancer prognosis, enabling complex integration of multimodal signals. 

In terms of treatment response and disease classification, LOMIA-T, a longitudinal Transformer framework was developed, which used cross-attention to model temporal changes in esophageal cancer images \cite{sun2024lomia}. In the field of cancer classification, Borah et al. proposed DCAT, a dual cross-attention fusion model that can enhance uncertainty-aware disease classification in radiological images \cite{borah2025dcat}. Sun et al. \cite{sun2023cross} designed a multi-branch CNN with a cross-attention module to classify breast cancer molecular subtypes using DCE-MRI, demonstrating the potential of attention-guided multi-branch architectures. Additionally, LASSO-MOGAT \cite{alharbi2024lasso} is a method that applies graph attention technology to integrate multi-omics data for cancer subtype classification, further revealing the practicality of attention mechanisms in high-dimensional biological data fusion. In thoracic disease classification, Ma et al. \cite{ma2019multi} implemented a cross-attention network to improve multi-label classification of chest X-rays. In brain disease, a selective cross-attention module using the Vision Transformer was developed to enhance brain tumor classification \cite{khaniki2024brain}. Similar ideas have been extended to fuse imaging and clinical data in recent years. For example, Amador et al. \cite{amador2025cross} used cross-attention to combine 4D CTP images with clinical metadata for functional stroke outcome prediction. And Verma et al. \cite{verma2024cross}  proposed a cross-attention fusion model that integrated omics, imaging, and clinical data to predict the prognosis of 130 lung cancer patients, demonstrating the flexibility of this mechanism in dealing with limited multimodal datasets.

Despite promising results, most medical imaging studies still target classification or prognosis on large oncology, stroke, or neurodegeneration cohorts. By contrast, musculoskeletal applications—such as femoral neck BMD regression—remain underexplored, and cross-attention has rarely been applied to fine-grained continuous regression in this setting. Notably, prior X-ray-based BMD studies \cite{feng2023deep, golestan2023approximating, ho2021application, hsieh2021automated, sato2022deep, wang2022lumbar} typically lack rich clinical metadata, which can substantially improve regression performance and enable clinical factor-level interpretability, as summarized in Table~\ref{tab:feature_matrix}. The table provides a comparative overview of existing approaches, highlighting the backbone models, input modalities, prediction tasks (BMD regression and T-score classification), feature fusion strategies, and interpretability mechanisms employed in these studies. To address these gaps, we developed XAttn-BMD, a cross-attention fusion model that integrates hip X-rays with structured clinical metadata. The method is detailed in Section~\ref{method}, and comparative results with these prior methods on our dataset are presented in Section~\ref{Comparative evaluation on our cohort}.

\begingroup
\renewcommand{\xmark}{\phantom{\cmark}}%
\begin{table*}[t]
\centering
\caption{Feature matrix of recent X-ray–based studies on BMD estimation and screening (binary classification). A \cmark\ denotes that the study used the feature; a blank denotes absence.}
\label{tab:feature_matrix}

\setlength{\tabcolsep}{4pt}
\renewcommand{\arraystretch}{1.12}
\setlength{\extrarowheight}{3.0ex}
\footnotesize

\begin{adjustbox}{max width=\linewidth}
\begin{tabular}{%
  C{2.2cm}  
  C{1.6cm}  
  C{1.8cm}  
  C{1.8cm}  
  C{2.0cm}  
  C{1.0cm}  
  C{2.2cm}  
  C{1.6cm}  
  C{2.0cm}  
  C{1.2cm}  
  C{1.2cm}  
  C{2.2cm}  
}
\toprule
\multicolumn{2}{c}{} &
\multicolumn{2}{c}{\textbf{Model Used}} &
\multicolumn{3}{c}{\textbf{Inputs}} &
\multicolumn{2}{c}{\textbf{Tasks}} &
\multicolumn{2}{c}{\textbf{Feature Fusion}} &
{} \\ 
\cmidrule(lr){3-4}\cmidrule(lr){5-7}\cmidrule(lr){8-9}\cmidrule(lr){10-11} 
\textbf{Study (Year)} & \textbf{Sample Size} & Model & Parameter \newline Count(M)$^1$  &
X-rays (location) & Metadata & Rich \newline metadata$^2$  &
BMD \newline Regression & T-score classification &
Cross-attention & Late concat &
\textbf{Interpretable}$^4$  \\ 
\midrule
Feng (2023)\cite{feng2023deep} & 139 & VGG16 DenseNet121 ResNet50 & $\approx 138.36\,\mathrm{M}$  $\approx 7.98\,\mathrm{M}$  $\approx 25.56\,\mathrm{M}$ & hip & \xmark & \xmark & \xmark & \cmark & \xmark & \xmark & \xmark \\
Golestan (2023)\cite{golestan2023approximating} & 533  & InceptionV3 & $\approx 24.71\,\mathrm{M}$ & hand, knee, pelvis & \cmark & \xmark & \cmark & \cmark & \xmark & \cmark & \xmark \\
Ho (2021)\cite{ho2021application} & 3,472   & ResNet18 & $\approx 11.25\,\mathrm{M}$ & pelvis  & \xmark & \xmark & \cmark & \cmark & \xmark & \xmark & \xmark \\
Hsieh (2021)\cite{hsieh2021automated}$^3$ & 5,164 18,175   & VGG16  ResNet34 &  $\approx 21.41\,\mathrm{M}$  $\approx 21.79\,\mathrm{M}$ & hip, lumbar & \cmark & \xmark & \cmark & \cmark & \xmark & \cmark & \xmark \\
Sato (2022)\cite{sato2022deep}   & 17,899   & ResNet50 & $\approx 24.58\,\mathrm{M}$  & chest  & \cmark & \xmark & \cmark & \cmark & \xmark & \cmark & \xmark \\
Wang (2022)\cite{wang2022lumbar} & 13,719   & VGG16 transformer & =24.93M & chest  & \xmark & \xmark & \cmark & \cmark & \xmark & \xmark & \cmark \\
XAttn-BMD \textbf{(Ours)}      & 233    & ResNet34 cross-attention & =23.32M & hip   & \cmark & \cmark & \cmark & \cmark & \cmark & \cmark & \cmark \\
\bottomrule
\end{tabular}

\end{adjustbox}
\begin{tablenotes}
  \tiny
  \item[1] $^1$ Parameter Count (M). "=" denotes values reported by the cited paper. "$\approx$" denotes our estimate based on the architecture described in the cited study and its canonical ImageNet implementation (3-channel input). For InceptionV3 in \cite{golestan2023approximating}, we assume text{aux\_logits}=False. Minor differences in library settings may result in slight variations that do not impact comparisons.
  \item[2] $^2$ "Rich metadata" denotes $\geq 5$ variables (e.g., age, sex, BMI, weight, height, smoking, alcohol, physical activity status).
  \item[3] $^3$ In \cite{hsieh2021automated}, the VGG16 model was used for hip X-rays, and ResNet34 was used for lumbar X-rays.
  \item[4] $^4$ The interpretability in Wang~(2022)\cite{wang2022lumbar} is at the image spatial level, and for XAttn-BMD, it is at the clinical metadata field level.
\end{tablenotes}
\end{table*}
\endgroup

\section{Materials and Method}
\subsection{Dataset and Preprocessing}
\begin{figure*}
    \centering
    \includegraphics[width=0.75\linewidth]{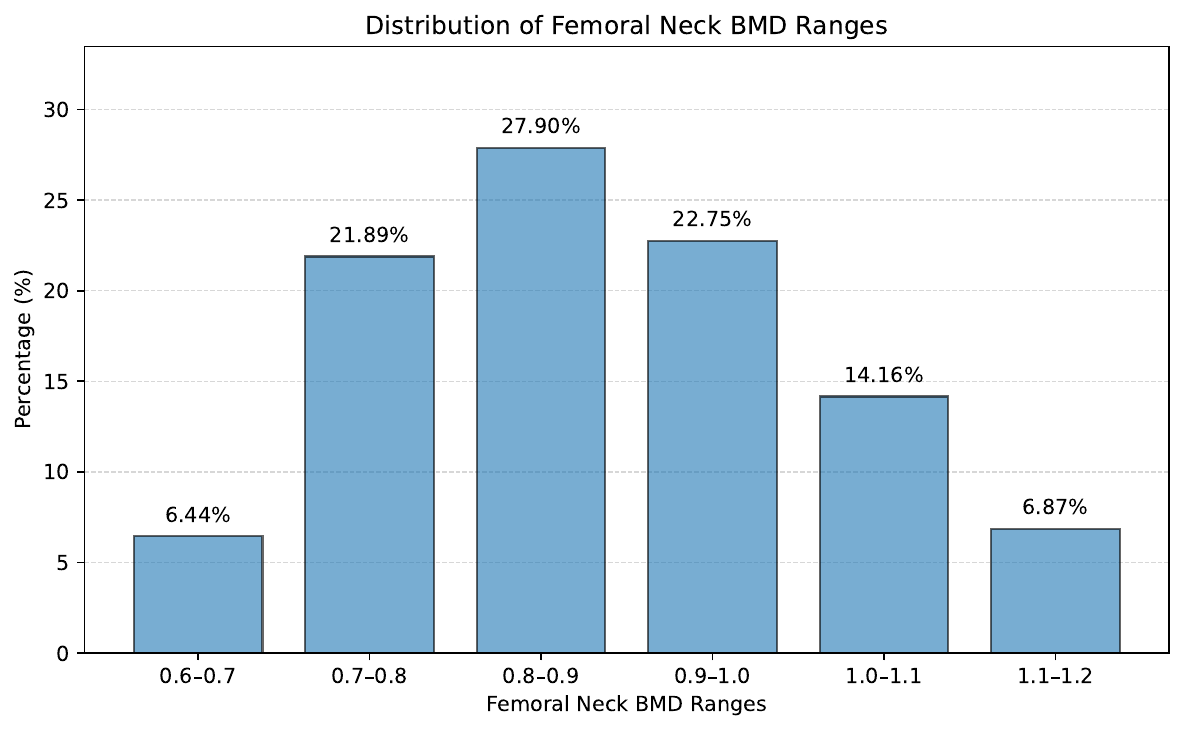}
    \caption{Distribution of femoral neck BMD}
    \label{fig:image-distribution}
\end{figure*}

The dataset used in this study is derived from the Hertfordshire Cohort Study \cite{syddall2019hertfordshire}, a community-dwelling cohort of older adults drawn from the general population and not selected on the basis of bone health. We used hip X-ray images for the estimation of femoral neck BMD. Each sample comprises a paired hip X-ray image and structured health-related metadata. In total, 233 hip X-ray images and their corresponding metadata entries were utilized for model training and evaluation, covering femoral neck BMD values ranging from 0.623 to 1.189 g/cm$^2$ (mean: 0.889~g/cm$^2$, std: 0.130~g/cm$^2$). The age of participants ranges from 71.6 to 80.6 years, reflecting an elderly population relevant to osteoporosis research. Images and metadata were preprocessed before training with the following steps, respectively. It is noteworthy that the dataset is imbalanced in terms of BMD distribution, with relatively fewer samples at the lower and upper extremes. The detailed distribution of BMD values is illustrated in Figure~\ref{fig:image-distribution}.

Additionally, Table~\ref{tab:metadata_summary} summarizes the structured metadata variables included in this study. These variables encompass a range of demographic, anthropometric, lifestyle, and clinical factors that are known to influence bone health and BMD. For each variable, the type, summary statistics (mean $\pm$ standard deviation for numerical variables, count and percentage for categorical variables), and measurement units are provided.

\begin{table*}[ht]
\centering
\footnotesize
\begin{threeparttable}
\caption{Summary of Structured Metadata Variables Used in the Study}
\label{tab:metadata_summary}
\begin{tabular}{llll}
\hline
\textbf{Variable (name in dataset)} & \textbf{Unit} & \textbf{Type} & \textbf{Distribution} \\
\hline
Age at x-ray (agexray)    & year   & Numerical   & $75.45 \pm 2.58$ \\
Age at BMD scan (hbsage)   & year   & Numerical   & $76.09 \pm 2.62$ \\
Height (epht)  & cm    & Numerical   & $166.17 \pm 8.79$ \\
Weight (epwt)  & kg  & Numerical   & $77.13 \pm 12.24$ \\
Body Mass Index (epbmi)  & kg/m$^2$  & Numerical   & $27.93 \pm 3.98$ \\
Alcohol consumption (epalunit)  & units/week     & Numerical   & $6.80 \pm 10.16$ \\
Physical activity in last 2 weeks (eptotact)  & mins/day  & Numerical   & $225.11 \pm 121.68$ \\
Diet quality score (epprddiet24)   & --    & Numerical   & $0.13 \pm 1.53$ \\
Sex (absex)    & --   & Categorical & \begin{tabular}[t]{@{}l@{}} Male: 119 (51.1\%) \\ Female: 114 (48.9\%) \end{tabular} \\
Smoking status (epsmkstat)   & --   & Categorical & \begin{tabular}[t]{@{}l@{}} Never: 121 (51.9\%) \\ Ex: 107 (45.9\%) \\ Current: 5 (2.1\%) \end{tabular} \\
\hline
\end{tabular}
\begin{tablenotes}
      \scriptsize
      \item[1] "--" indicates no unit for this variable.
      \item[2] The metadata variable names in parentheses (e.g., agexray, epbmi) correspond to the x-axis variable names in Figure~\ref{fig:image-field_attention_fold3}.
\end{tablenotes}
\end{threeparttable}
\end{table*}

\subsubsection{Image Processing}
\begin{enumerate}
    \item Region of interest segmentation:

Bone mineral density is usually measured in fixed areas, but there are many other areas in addition to the characteristic area being measured in hip X-ray images. These areas contain a lot of features that are not required for BMD measurement. Therefore, to better capture features relevant to femoral neck BMD, the hip X-ray images are preprocessed by extracting the bilateral femoral neck regions as the regions of interest for subsequent BMD estimation. 
    
BoneFinder \cite{lindner2013fully}, as a shape-based anatomical modeling tool, is employed to perform accurate segmentation of the femoral neck regions. Subsequently, to reduce redundant black background pixels and enhance the effectiveness of feature extraction, the two segmented femoral neck regions were merged. The segmentation and merging process is shown in Figure~\ref{fig:image-segment-process}.
\begin{figure*}
    \centering
    \includegraphics[width=1\linewidth]{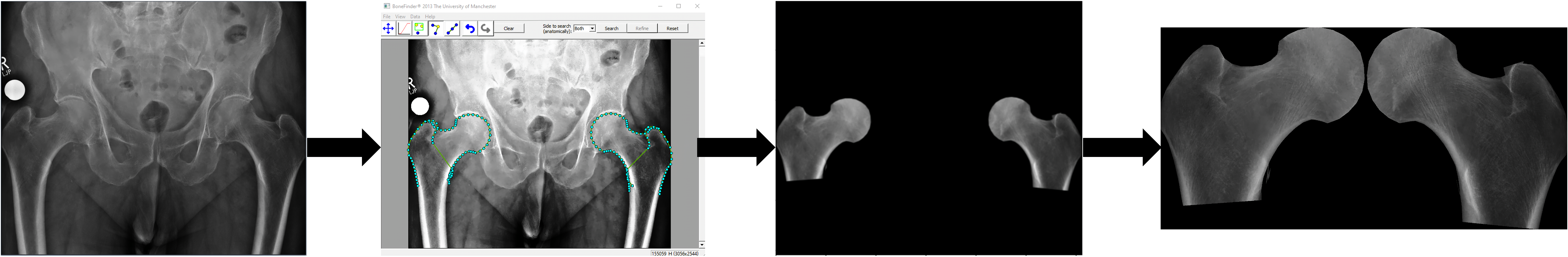}
    \caption{Process of segmenting the hip X-ray image with Bone Finder and Merging}
    \label{fig:image-segment-process}
\end{figure*}

    \item Image augmentation: 
 
To address the imbalanced distribution of BMD values and reduce overfitting, image augmentations were applied exclusively to the training set. The number of augmentations for each image was determined by its associated BMD value. Specifically, samples in underrepresented ranges, such as low BMD (0.6–0.7) and high BMD (1.1–1.2), were augmented the most frequently, while samples near the central distribution (e.g., 0.9) received fewer augmentations. This strategy increased the representation of clinically important but infrequent cases.

Since the images represent a cropped region of interest (ROI) around the femoral neck, where pixel intensities have anatomical significance related to bone mineral density, we carefully selected augmentation techniques to preserve the intensity values. We applied geometry-based transformations, including horizontal flipping, affine translation, small-angle rotation, and shearing, as well as the introduction of Gaussian noise for augmentation. These augmentation operations preserve image pixels while introducing enough variability and simulate image differences caused by different imaging devices. The details of the augmentation operations are shown in Table~\ref{tab:training-augmentations}. For each image, 1 to 3 transformations were randomly selected to enhance diversity.
\begin{table*}[htbp]
\centering
\footnotesize
\caption{Random data augmentation operations used during training}
\begin{tabular}{ccc}
\hline
\textbf{Augmentation} & \textbf{Description / Parameters} & \textbf{Probability} \\
\hline
GaussNoise & moderate noise level, per-channel & p=0.5 \\
HorizontalFlip & horizontal axis flip & p=0.5 \\
Affine (scale) & scale range: 0.95–1.05 & p=0.5 \\
Affine (translate) & translation range: ±3\% & p=0.5 \\
Affine (rotate) & rotation range: ±10° & p=0.5 \\
Affine (shear) & shear range: ±3° & p=0.5 \\
RandomBrightnessContrast & brightness\_limit:0.1, contrast\_limit:0.1 & p=0.5 \\
\hline
\end{tabular}
\label{tab:training-augmentations}
\end{table*}

No augmentation was applied to the test set during training, ensuring a fair evaluation and consistency with clinical interpretation. Representative examples of augmented images (resized to 224×224 to meet ResNet34 input requirements) are shown in Figure~\ref{fig:image-Examples-of-image-augmentation-and-resizing-to-224×224}.
\begin{figure*}
    \centering
    \includegraphics[width=1\linewidth]{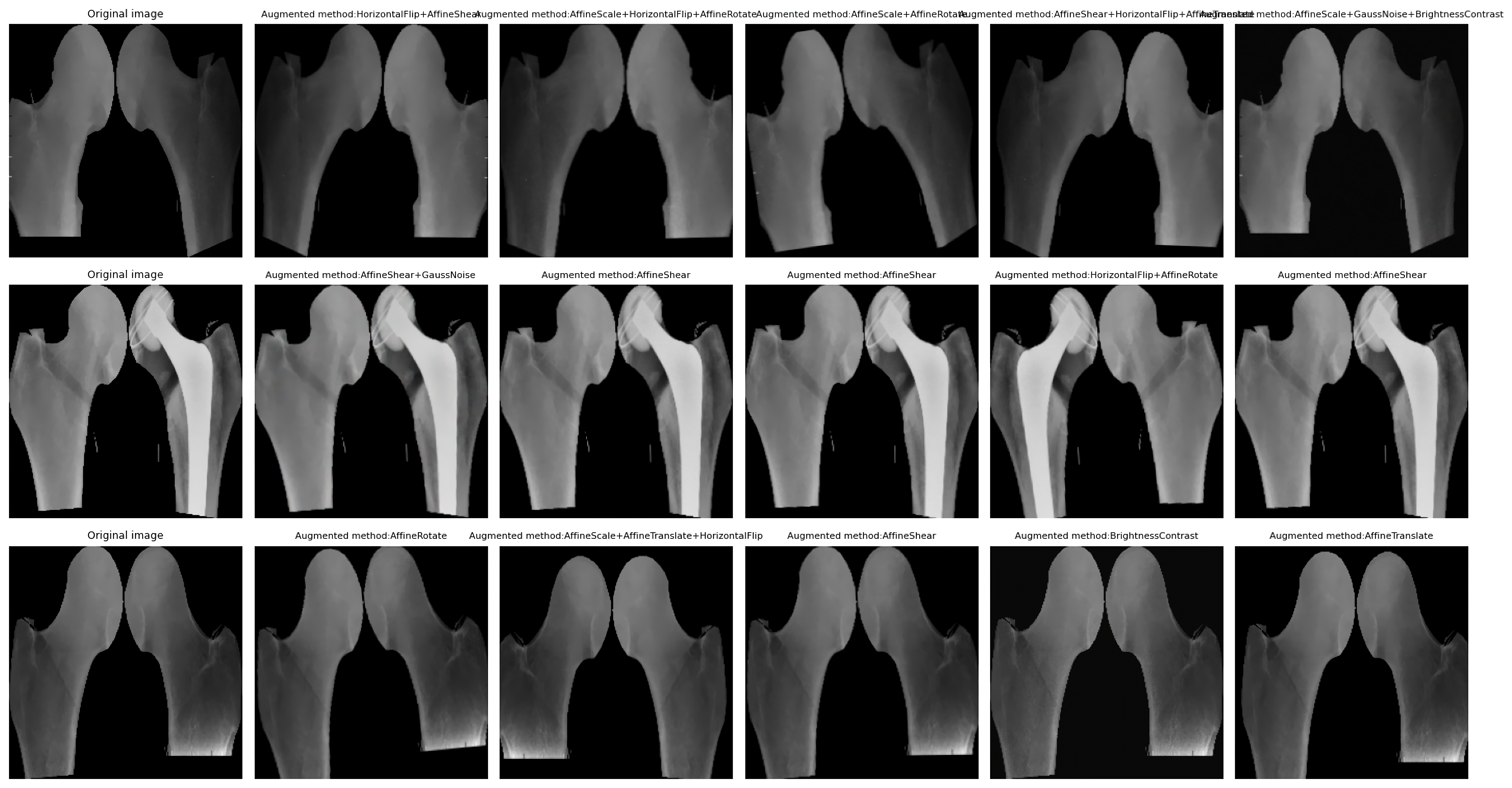}
    \caption{Examples of image augmentation and resizing to 224×224}
    \label{fig:image-Examples-of-image-augmentation-and-resizing-to-224×224}
\end{figure*}

\end{enumerate}
\subsubsection{Metadata Processing}

The structured metadata comprises both numerical and categorical features. Numerical variables, such as age, height, weight, and physical activity levels, were standardized using the StandardScaler to achieve zero mean and unit variance. Categorical variables, including sex and smoking status, were transformed via one-hot encoding to produce binary feature vectors compatible with neural network input. The final metadata representation was formed by concatenating the standardized numerical features and the one-hot encoded categorical vectors, and then cast to float32 to ensure compatibility with the model input.
\subsection{Methods}
\label{method}
\subsubsection{Overview of the Proposed Model}
\begin{figure*}
    \centering
    \includegraphics[width=1\linewidth]{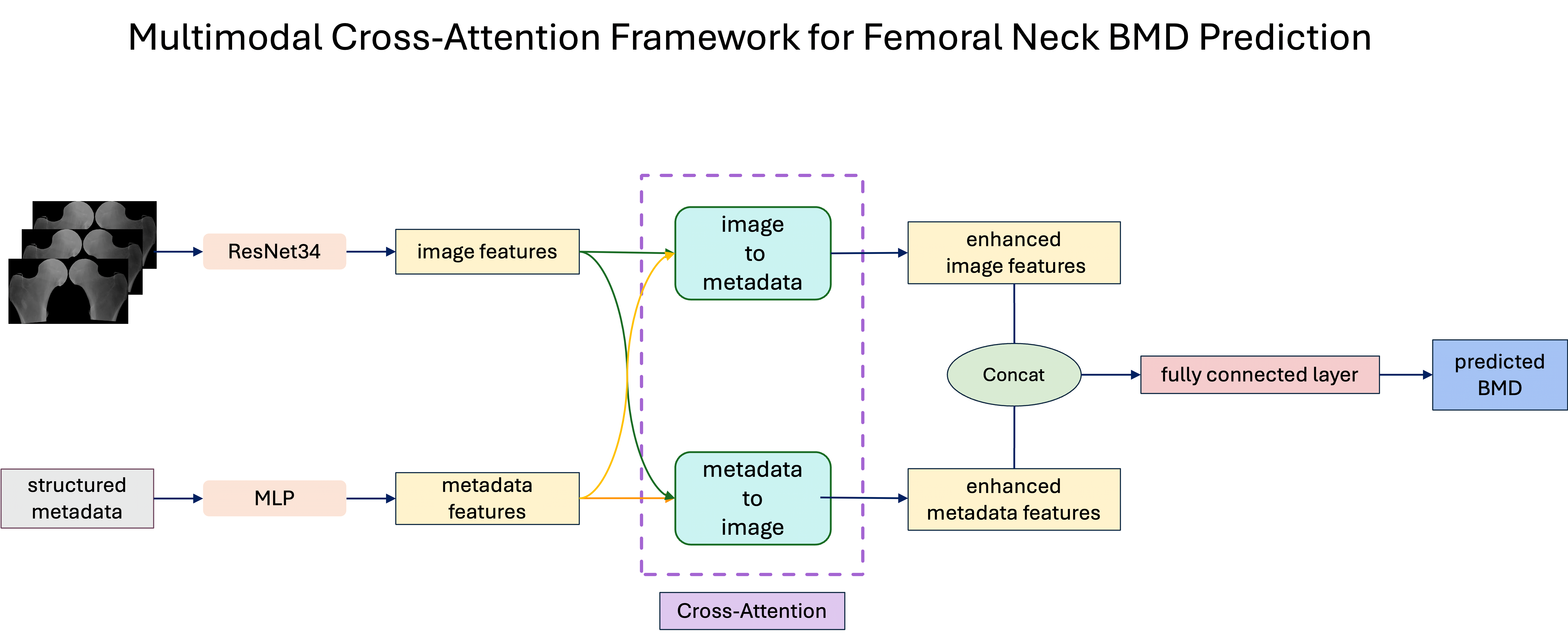}
    \caption{Overview of \textbf{XAttn-BMD}. Hip X-rays and structured clinical metadata are encoded and fused via bidirectional cross-attention to predict femoral-neck BMD. \(C\) denotes late concatenation of the enhanced modality features.}

    \label{fig:image-overview}
\end{figure*}

The overall framework of XAttn-BMD is shown in Figure~\ref{fig:image-overview}. In this framework, image features are extracted using a ResNet34 backbone pre-trained on ImageNet, providing a rich representation of the visual input. Meanwhile, the structured metadata is encoded through an MLP, which transforms tabular data into a dense feature embedding. The extracted image and metadata features are then jointly processed through a vector-level cross-attention mechanism, enabling the model to capture high-level interactions between visual and structured information. The cross-attention module consists of two complementary branches: the image-to-metadata attention branch, which enables the model to query relevant metadata features conditioned on image representations. The metadata-to-image attention branch allows the model to query corresponding image regions based on metadata information. The enhanced features obtained from the image-to-metadata and metadata-to-image attention branches are concatenated and then input into a fully connected layer to perform the final estimation of femoral neck bone mineral density. 

Compared to directly concatenating image features and metadata features, this bidirectional attention facilitates a more informative and interpretable interaction of multimodal features for the downstream BMD regression.
\subsubsection{Feature Extractor}
To effectively represent both modalities, two separate extractors are employed for the two modalities, respectively:
\begin{enumerate}
\item \textbf{Image feature extractor}

To extract features from the input images, a ResNet34 model is employed. We choose ResNet34 for its strong accuracy-efficiency trade-off on our relatively small dataset: it is deeper than ResNet18 yet lighter than ResNet50, which helps mitigate overfitting while keeping computation moderate. The final classification layer of ResNet34 is removed, and a global average pooling layer is applied to the final convolutional feature map, obtaining a 512-dimensional image feature vector. This vector is then passed through a fully connected projection layer with ReLU activation to produce the final image embedding used for later cross-modal fusion.
\item \textbf{Metadata feature extractor} 

The structured metadata, consisting of normalized continuous features and one-hot encoded categorical features, is processed using a two-layer MLP. The first layer projects the input into a 128-dimensional hidden space, followed by a ReLU activation and dropout regularization. The second layer further maps the features to a 64-dimensional latent embedding, which is also used as the input to the cross-attention mechanism for multimodal feature fusion.
\end{enumerate}
\subsubsection{Bidirectional Vector-level Cross-attention Module}
This module employs a lightweight multi-layer, multi-head cross-attention mechanism to integrate features between two modalities: a query modality and a key-value modality. It consists of three sequential attention layers, each containing four parallel attention heads. In every layer, both the query and key-value features are linearly projected into queries (Q), keys (K), and values (V), and the projected vectors are split across heads to compute attention.

To enhance the representational capacity of the key-value modality, we introduce a key-value updater, implemented as a small feed-forward network. This updater is applied after each layer to refine the key and value features, and operates on the key-value features, independent of the query modality or attention outputs. It ensures that the key and value embeddings can evolve across layers while remaining computationally efficient.

Additionally, a layer-wise head-shared mechanism is employed to modulate the relative importance of each attention head. Specifically, a learnable scalar weight is shared across heads within each layer, enabling the model to fuse multi-head information adaptively. The updated query and key-value representations are iteratively refined through the three cross-attention layers, and the final query embedding is used as the fused representation for downstream BMD regression.

\begin{figure*}
    \centering
    \includegraphics[width=1\linewidth]{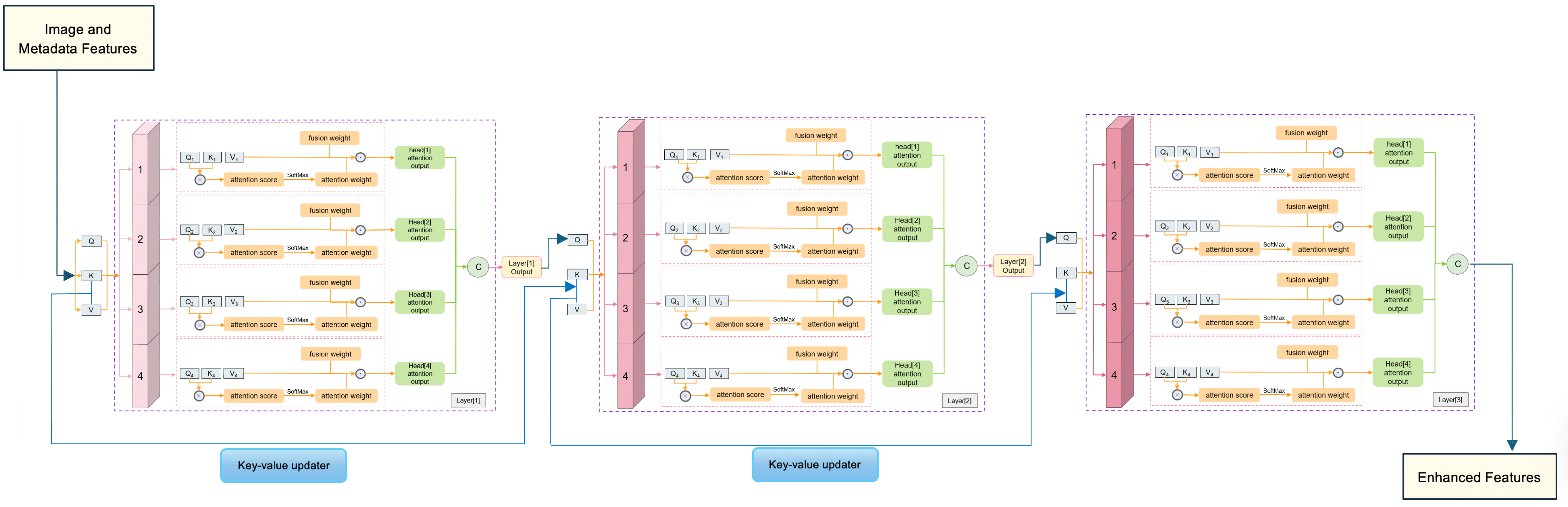}
    \caption{The architecture of the cross‐attention module. \(C\) denotes concatenation across attention heads, \(X\) denotes matrix multiplication, and “\(\odot\)” denotes element‐wise multiply.}
    \label{fig:image-cross-attention-architecture}
\end{figure*}

Image-to-metadata and metadata-to-image branches in the Cross-Attention module have the same architecture, shown in Figure~\ref{fig:image-cross-attention-architecture}. 

The module processes the input image and metadata features and generates the enhanced features through the following steps:
\\a) Linear projection: 

Before computing attention, the image and metadata features are linearly projected into a shared representation space to produce the Query (Q), Key (K), and Value (V) vectors. This is achieved using three independent linear transformations:
\begin{equation}
Q = X W_Q,\quad K = Y W_K,\quad V = Y W_V
\end{equation}
where \(X\) denotes the query input (e.g., image features in the image-to-metadata branch, and metadata features in the metadata-to-image branch), \(Y\) denotes the key-value input (e.g., metadata features in image-to-metadata, image features in metadata-to-image), and \(W_Q, W_K, W_V\) are projection matrices.
\\b) Attention score: 

The attention score measures how much focus the model should place on each key when generating the output for a given query. It quantifies the relevance of each Query-Key pair by computing their scaled dot product. A higher score indicates that the corresponding value is more informative for the current query. Let \( Q_i \) and \( K_i \) be the projected query and key vectors for the \( i \)-th position, respectively. The attention score is calculated as:

\begin{equation}
\text{AttentionScore}(Q_i, K_i) = \frac{Q_i \cdot K_i}{\sqrt{d_k}}
\end{equation}
where \( d_k \) is the dimension of the key vectors. The scaling factor \( \sqrt{d_k} \) helps stabilize gradients during training by preventing large dot-product values.
\\c) Attention weight:

The attention scores are normalized into probability distributions using a SoftMax function, enabling the aggregation of value vectors. This normalization assigns higher weights to more relevant key-value pairs, thereby guiding the final representation to focus on information-rich elements.
\begin{equation}
\text{AttentionWeight} = \text{SoftMax}(\text{AttentionScore})
\end{equation}
d) Adaptive fusion (fusion weight):

To balance model expressiveness and generalization on a limited-size dataset, we adopt an adaptive fusion strategy based on layer-wise, head-shared fusion weights. In each attention layer, a learnable scalar weight is uniformly applied across all attention heads to scale the output of the value vectors. This scalar serves as a layer-specific global scaling factor, modulating the magnitude of the attention response.

By sharing fusion weights within each layer but learning them independently across layers, the model maintains intra-layer consistency while enabling inter-layer flexibility. This design helps reduce parameter redundancy while allowing the network to dynamically control the relative contributions of different layers to the final fused representation. As a result, the approach enhances the stability and adaptability of multimodal learning.
\\e) Head output:

After computing the attention weights in each head, the corresponding value vectors are scaled accordingly to produce head-specific outputs. A learnable fusion weight, shared across all heads within the same layer, further modulates this process. This design allows each attention head to focus on different aspects of the feature space, thereby enhancing the model’s capacity to capture diverse patterns across modalities.

Formally, the output of the $h$-th head in layer $l$ is computed as:
\begin{equation}
\text{HeadOutput}^{(h)} = \text{AttentionWeight}^{(h)} \cdot V^{(h)} \cdot \text{FusionWeight}^{(l)}
\end{equation}
where $\text{AttentionWeight}^{(h)}$ represents the normalized attention scores for head $h$, $V^{(h)}$ denotes the corresponding value vectors, and $\text{FusionWeight}^{(l)}$ is the layer-specific learnable scalar applied uniformly across all heads in layer $l$, where $l$ ranges from 1 to 3 and $h$ ranges from 1 to 4.
\\f) Layer output:

After computing the attention-weighted value representations from all heads, which are referred to as head outputs, these are concatenated and passed through a layer-specific linear projection. This projection transforms the aggregated multi-head representation into the desired output dimensionality and serves as the final output of the current attention layer. For the $l$-th layer, the output can be formulated as follows:
\begin{equation}
\text{LayerOutput}^{(l)} = W^{(l)} \cdot \operatorname{Concat}(\text{HeadOutput}^{(h)})
\end{equation}
where $W^{(l)}$ denotes the learnable weight matrix of the linear projection for layer $l$, and the concatenation is performed over the outputs of all heads within that layer.
\\g) key-value updater (layer-wise refinement of key and value features): 

To progressively refine the key and value features across attention layers, each layer incorporates a lightweight feed-forward updater module. This updater consists of a Layer Normalization, a linear transformation, a GELU activation, and a dropout layer. At the \( l \)-th layer, the updater transforms the input key-value feature \( \mathbf{Y}^{(l)} \) into an update vector, which is then added residually to the feature from the previous layer with a scaling factor of 0.5, to obtain the refined key-value representation for the next layer. The original key-value features come from the feature extractors. The process can be formulated as follows:
\begin{equation}
\mathrm{Updater}^{(l)}(\mathbf{Y}) = \mathrm{Dropout}(\mathrm{GELU}(\mathrm{Linear}(\mathrm{LayerNorm}(\mathbf{Y}))))
\end{equation}
\begin{equation}
\mathbf{Y}^{(l+1)} = \mathbf{Y}^{(l)} + 0.5*\mathrm{Updater}^{(l)}(\mathbf{Y}^{(l)})
\end{equation}
This residual updating mechanism enables the model to progressively enrich key and value representations at each layer while preserving training stability and model expressiveness.
\\h) Enhanced feature fusion:

After applying the cross-attention mechanism, the original unimodal features are enriched by incorporating complementary information from the other modality. Specifically, image features are enhanced via interactions with metadata features (image-to-metadata attention), and metadata features are concurrently enhanced through interactions with image features (metadata-to-image attention). These enriched representations are referred to as enhanced features, as they integrate contextual cues across modalities. The outputs from both attention directions, enhanced image features and enhanced metadata features, are concatenated to form a unified multimodal representation, which is subsequently passed through a fully connected layer to predict femoral neck BMD. Formally, the process is expressed as:

\begin{align}
\mathbf{F}_{\text{img-enhanced}} &= \text{CrossAttn}_{\text{img} \rightarrow \text{meta}}(\mathbf{F}_{\text{img}}, \mathbf{F}_{\text{meta}}) \\
\mathbf{F}_{\text{meta-enhanced}} &= \text{CrossAttn}_{\text{meta} \rightarrow \text{img}}(\mathbf{F}_{\text{meta}}, \mathbf{F}_{\text{img}}) \\
\mathbf{F}_{\text{enhanced}} &= \text{Concat}(\mathbf{F}_{\text{img-enhanced}}, \mathbf{F}_{\text{meta-enhanced}})
\end{align}
where \( \mathbf{F}_{\text{img-enhanced}} \in \mathbb{R}^{B \times d_i} \) and \( \mathbf{F}_{\text{meta-enhanced}} \in \mathbb{R}^{B \times d_m} \) denote the enhanced image and metadata features, respectively, and \( B \) is the batch size. The final concatenated feature \( \mathbf{F}_{\text{enhanced}} \in \mathbb{R}^{B \times (d_i + d_m)} \) is used as input to the regression head for BMD estimation.

\subsubsection{Weighted Smooth L1 Loss}

In this dataset, the distribution of femoral neck BMD values is imbalanced, with the majority of samples concentrated in the clinically normal range, and only a small proportion corresponding to low BMD conditions such as osteopenia or osteoporosis. This imbalance presents a unique challenge: although most samples contribute to training, the clinically significant cases---those at the low end of the BMD spectrum---are underrepresented. Effective identification and regression of these rare cases is critical for early diagnosis and intervention.

To address this challenge, we propose a Weighted Smooth L1 Loss, which extends the standard Smooth L1 loss (Huber Loss) by incorporating an adaptive weighting scheme that emphasizes underrepresented samples. Specifically, samples whose ground truth BMD values deviate further from a predefined center are assigned larger weights, encouraging the model to focus more on boundary cases. The per-sample loss is defined as follows:

\begin{equation}
\mathcal{L}_{\text{WeightedSmoothL1Loss}} = w_i \cdot \mathcal{L}_{\text{HuberLoss}}(y_i, \hat{y}_i)
\end{equation}
\begin{equation}
w_i = 1 + \lambda \cdot |y_i - c|
\end{equation}

where \( y_i \) and \( \hat{y}_i \) denote the ground truth and predicted BMD values, respectively, \( c \) is the predefined center point set to \(0.9~\mathrm{g/cm^2}\), and \( \lambda \) is a hyperparameter that controls the rate at which the weight increases with deviation. The function \( \mathcal{L}_{\text{HuberLoss}} \) denotes the standard Huber loss with transition parameter \( \beta = 1.0 \), ensuring a balance between outlier resistance and smooth gradients.

During training, we minimize the mean over a mini-batch of size \(N\):
\begin{equation}
\mathcal{L} \;=\; \frac{1}{N}\sum_{i=1}^N \mathcal{L}_{\text{WeightedSmoothL1Loss}}^{(i)}
\end{equation}

Compared to standard Mean Squared Error (MSE) or unweighted Huber loss (Smooth L1 Loss), the proposed Weighted Smooth L1 Loss explicitly increases the influence of clinically significant, low-frequency cases on gradient updates. This design leads to improved regression performance on rare but critical BMD values, while retaining stability for the majority distribution.
\subsubsection{Stratified 10-fold Cross-validation}
To ensure balanced evaluation across varying BMD ranges, stratified 10-fold cross-validation was employed based on discretized BMD intervals. The continuous BMD values of the femoral neck were first partitioned into equally spaced bins ranging from 0.6 to 1.2 with a step size of 0.1, reflecting clinically significant ranges. Each sample was then assigned a bin label, and a Stratified Fold strategy was applied using these bin labels as stratification targets. This stratification strategy helps reduce performance variance across folds by maintaining consistent BMD distribution, especially important in this imbalanced regression task.
\subsubsection{Ablation Experiments}
To better understand the contribution of each component in the proposed model, we conducted a comprehensive ablation study, including modality comparison, architectural variants, and attention mechanism validation. The experiments are grouped into the following aspects.
\begin{enumerate}
    \item Modality contribution: To investigate the impact of different modalities, we evaluated the model using image features only, metadata features only, and both modalities combined. This can reveal the individual and joint predictive power of image and metadata features for estimating femoral neck BMD.
    \item Fusion strategy comparison: The proposed cross-attention-based fusion model is compared with a baseline model that directly concatenates image and metadata features. This comparison can evaluate the effectiveness of explicitly modeling cross-modal interactions over simple early fusion in predicting femoral neck BMD.
    \item Dual-direction cross-attention validation: To validate the benefit of bidirectional cross-attention, the model is evaluated under three configurations: using only image-to-metadata attention, only metadata-to-image attention, and the full bidirectional setup. This analysis can highlight the contribution of mutual enhancement between modalities to the overall performance.
\end{enumerate}
\subsubsection{Evaluation Metrics}
To quantitatively assess the performance of the regression models, three standard metrics are employed on the test set to evaluate the model: Mean Squared Error (MSE), Mean Absolute Error (MAE), and the coefficient of determination ($R^2$ score).
\begin{enumerate}
  \item \textbf{Mean Squared Error (MSE)}: Measures the average of the squared differences between predicted and actual values and penalizes larger errors more severely, making it sensitive to outliers. Lower MSE indicates better overall model fit. This property allows MSE to emphasize large deviations, which is beneficial for capturing overall regression consistency and detecting significant errors that may occur in critical clinical cases.

  \item \textbf{Mean Absolute Error (MAE)}: Calculate the average absolute difference between the estimates and the ground truth. Unlike MSE, MAE treats all errors equally and is less sensitive to the distribution tails. In this dataset, a subset of tail-end samples corresponds to clinically significant low BMD values, which require tolerance to tail-end samples during training.

  \item \textbf{$R^2$ score}: Indicates proportion of the variance in the target variable that is predictable from the input features. An $R^2$ score value closer to 1 suggests a better fit to the model, while values below 0 indicate that the model performs worse than a simple mean baseline. $R^2$ score is particularly useful to evaluate how well the model explains the overall variance in BMD across different patient profiles, which complements point-wise metrics such as MSE and MAE by providing a global measure of the fit of the model.
\end{enumerate}

Beyond regression performance, the model outputs were also assessed in a binary classification framework to evaluate their utility for clinical screening. A clinical threshold of femoral neck BMD was used to dichotomize estimates and ground truth, and binary classification metrics such as ROC-AUC, precision, recall and F1-score were reported to assess the model's screening performance of detecting low femoral neck BMD subjects.

\section{Results}
\subsection{Implementation Details}
All models, including ablation models and the proposed model, were trained using the Adam optimizer, with hyperparameters (e.g., learning rate, weight decay, number of epochs, and L1 regularization) individually tuned to achieve the best performance for each configuration in 10-fold cross-validation with a batch size of 32 for training and batch size of 8 for testing. For fair comparison, a series of experiments was conducted to identify the optimal settings for each model variant. Specifically, in the proposed cross-attention model, training was performed for 400 epochs using a learning rate of $1 \times 10^{-4}$, with a weight decay of $3 \times 10^{-5}$ and a slight L1 regularization term ($\lambda = 5 \times 10^{-7}$) to reduce the risk of overfitting. To further prevent overfitting, Dropout was applied at different stages of the model: 0.2 in the image projection head and the metadata MLP, 0.1 in the key-value updater, and 0.05 in the final regression fully connected layer. The scaling factor (hyperparameter $\lambda$) in Weighted Smooth L1 loss is set to 3.0. A mechanism similar to early stopping based on validation loss was adopted, and the checkpoint with the best validation performance was saved and used for final evaluation. Training was conducted on an NVIDIA A100 GPU node. Each epoch took approximately 4 seconds, resulting in a total training time of about 4.5 hours for the 10-fold cross-validation.
\subsection{Comparison of Loss Functions}
\label{Comparison-of-loss-functions}
To evaluate the effectiveness of the proposed Weighted Smooth L1 Loss, we compared it with two widely used regression loss functions: Mean Squared Error (MSE) and Huber loss (Smooth L1 loss) in the proposed model. Each loss function was independently employed during model training and validation to assess its impact on regression performance. For consistency, all models were evaluated on the test set using the MSE metric. This setup ensures a fair comparison across different training objectives while using a common evaluation criterion.

\begin{figure*}
    \centering
    \includegraphics[width=1.0\linewidth]{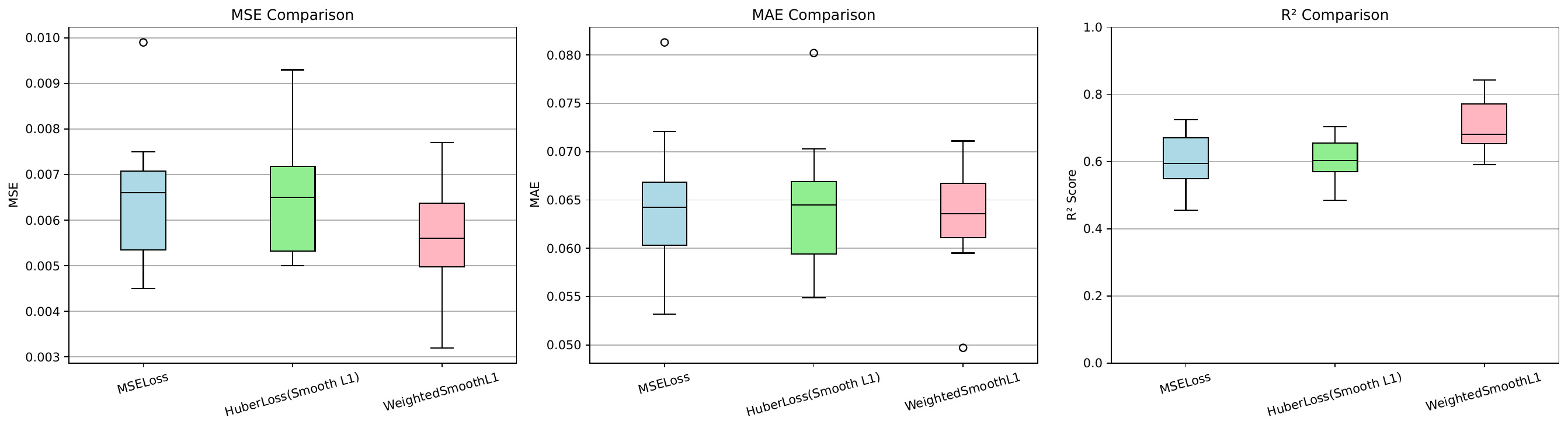}
    \caption{Comparison of regression loss functions across 10-fold CV using MSE, MAE, and$R^2$ metrics.}
    \label{fig:image-loss-compare}
\end{figure*}

As shown in the Figure~\ref{fig:image-loss-compare}, the proposed Weighted Smooth L1 loss function consistently outperforms MSELoss and HuberLoss (Smooth L1) on all evaluation metrics. Specifically, for the MSE and MAE metrics, the Weighted Smooth L1 loss function achieves the lowest median. It exhibits a narrower interquartile range, indicating better regression and cross-fold consistency. In terms of $R^2$, it achieves the highest median score with low variability, indicating its superior ability in capturing the variance of femoral neck BMD.

Table~\ref{tab:loss_comparison} shows that the Weighted Smooth~L1 loss improves BMD estimation reliability, with notable gains on under-represented BMD cases.

\begin{table*}[htbp]
\centering
\footnotesize
\caption{Performance comparison of different loss functions across 10-fold cross-validation. All results are obtained with the proposed model.}
\label{tab:loss_comparison}
\begin{tabular}{lccc}
\hline
\textbf{Loss Function} & \textbf{MSE} & \textbf{MAE} & $\mathbf{R^2}$ \\
\hline
Weighted Smooth L1 & 0.0055 $\pm$ 0.0012 & 0.0592 $\pm$ 0.0059 & 0.7011 $\pm$ 0.0802 \\
MSELoss            & 0.0065 $\pm$ 0.0014 & 0.0646 $\pm$ 0.0077 & 0.6043 $\pm$ 0.0829 \\
HuberLoss          & 0.0065 $\pm$ 0.0014 & 0.0644 $\pm$ 0.0069 & 0.6062 $\pm$ 0.0688 \\
\hline
\end{tabular}
\end{table*}

\subsection{Ablation Experiments Results}
\subsubsection{Modality Contribution Results}

Firstly, to assess the contribution of metadata in addition to imaging information, we compared the model performance using two different input settings: (1) image-only model, where the model was trained solely on hip X-ray images; (2) metadata-only model, the model only trained on health-related metadata; and (3) multimodal, where both hip X-ray images and structured metadata were utilized as inputs. Here, features from two modalities are concatenated directly, without using cross-attention to fuse. All experiments were conducted under the best training version and the same evaluation protocol to ensure comparability.

These results are summarized in Table~\ref{tab:modality_comparison}. The observed improvement in model performance after incorporating metadata in training demonstrates that metadata further enhances the predictive performance of the model by providing complementary information to the images. The structured health-related variables contribute clinical context—such as age, sex, and lifestyle factors—which help the model better account for individual variability in bone health. This highlights the value of integrating both image and metadata for more accurate BMD estimation in clinical applications.
\begin{table*}[htbp]
\centering
\footnotesize
\caption{Performance comparison between single modality and multimodal models.}
\label{tab:modality_comparison}
\begin{tabular}{cccc}
\hline
\textbf{Input Modality} & \textbf{MSE} & \textbf{MAE} & $\mathbf{R^2}$ \\
\hline
Image-only & 0.0074 $\pm$ 0.0013 & 0.0681 $\pm$ 0.0073 & 0.5528 $\pm$ 0.0897 \\
Metadata-only & 0.0124 $\pm$ 0.0033 & 0.0880 $\pm$ 0.0106 & 0.2537 $\pm$ 0.1747 \\
\makecell[c]{Image + Metadata\\\quad (concatenate fusion)} & 0.0066 $\pm$ 0.0013 & 0.0630 $\pm$ 0.0056 & 0.6022 $\pm$ 0.0798 \\
\hline
\end{tabular}
\end{table*}

\subsubsection{Fusion Strategy Comparison Results}
\label{sec:Fusion-strategy-comparison-results}
Having demonstrated the positive impact of adding metadata, to evaluate the effectiveness of the cross-attention modality fusion strategy, we compared the concatenate fusion method with the proposed cross-attention fusion mechanism. The advantage of cross-attention fusion is that it can dynamically weigh and align features from both modalities, allowing the model to focus on the most informative aspects of each input. In contrast, simple connections treat all features equally and may not capture subtle relationships or hierarchical dependencies.

As shown in Table~\ref{tab:fusion_comparison}, the cross-attention fusion strategy achieves significant improvements over the traditional concatenate fusion method for fusing features from different modalities. Specifically, the cross-attention-based model achieves lower mean square error and higher $R^2$ score while maintaining comparable mean absolute error. This suggests that cross-attention can more effectively integrate image features and clinical metadata, resulting in better estimation performance.

The superior results of cross-attention fusion highlight its ability to adaptively model complex interactions between modalities, rather than treating features as independent and equally important. In summary, these findings suggest that using the cross-attention mechanism is a more powerful and flexible multimodal learning strategy in the femoral neck BMD regression.
\begin{table*}[htbp]
\centering
\footnotesize
\caption{Comparison of concatenate and cross-attention fusion strategies.}
\label{tab:fusion_comparison}
\begin{tabular}{lccc}
\hline
\textbf{Fusion Strategy} & \textbf{MSE} & \textbf{MAE} & $\mathbf{R^2}$ \\
\hline
Concatenate Fusion   & 0.0066 $\pm$ 0.0013 & 0.0630 $\pm$ 0.0056 & 0.6022 $\pm$ 0.0798 \\
Cross-Attention Fusion & 0.0055 $\pm$ 0.0012 & 0.0592 $\pm$ 0.0059 & 0.7011 $\pm$ 0.0802 \\
\hline
\end{tabular}
\end{table*}

\subsubsection{Bidirectional Cross-attention Validation results}
\label{Bidirection-cross-attention-validation-results}

To further explore the impact of the bidirectional cross-attention mechanism, we evaluate the model performance under different strategies for fusing image features with metadata features. As shown in Table~\ref{tab:bidirectional_cross_attention_validation}, the fully bidirectional cross-attention model achieves the highest $R^2$ score of approximately 0.7011, outperforming the unidirectional and concatenated fusion strategies. Specifically, the model that only adopts the metadata-to-image (Metadata$\rightarrow$Image) attention mechanism achieves a relatively high $R^2$ of 0.5867, indicating that metadata can effectively guide attention to image features. In contrast, the model with the image-to-metadata (Image$\rightarrow$Metadata) attention mechanism produces a much lower $R^2$ of only 0.1582, indicating that the ability to query structured metadata based on visual features alone is limited. These results show that while metadata-to-image attention contributes more directly to the regression performance, the combination of the two directions in the bidirectional cross-attention mechanism enables the model to more effectively exploit the complementary information from the two modalities, resulting in more robust and general BMD estimation.

\begin{table*}[htbp]
\footnotesize
\centering
\caption{Comparison of different attention strategies for image and metadata features.}
\label{tab:bidirectional_cross_attention_validation}
\begin{tabular}{cccc}
\hline
\textbf{Attention Strategy} & \textbf{MSE} & \textbf{MAE} & $\mathbf{R^2}$ \\
\hline
Image$\rightarrow$Metadata    & 0.0138 $\pm$ 0.0039 & 0.0921$\pm$ 0.0153 & 0.1582 $\pm$ 0.2484 \\
Metadata$\rightarrow$Image   & 0.0068 $\pm$ 0.0008 & 0.0657 $\pm$ 0.0046 & 0.5867 $\pm$ 0.0479 \\
\makecell[c]{Bidirectional Cross-Attention\\\quad(proposed)} & 0.0055 $\pm$ 0.0012 & 0.0592 $\pm$ 0.0059 & 0.7011 $\pm$ 0.0802 \\
\hline
\end{tabular}
\end{table*}

\subsection{Performance of the Proposed Model}
\subsubsection{Cross-validation Performance}
The proposed model was evaluated using 10-fold cross-validation. As summarized in Table~\ref{tab:loss_comparison} (results of Weighted Smooth L1) in Section~\ref{Comparison-of-loss-functions}, Table~\ref{tab:fusion_comparison} in Section~\ref{sec:Fusion-strategy-comparison-results} (results of Cross-Attention Fusion), and Table~\ref{tab:bidirectional_cross_attention_validation} (results of Bidirectional Cross-Attention) in Section~\ref{Bidirection-cross-attention-validation-results}, the model achieved an average \textbf{mean squared error (MSE) of 0.0055 ± 0.0012, Mean Absolute Error (MAE) of 0.0592 ± 0.0059, and coefficient of determination ($R^2$ score) of 0.7011 ± 0.0802}. These results indicate that the model is capable of making an consistent estimation of femoral neck BMD. Using out-of-fold predictions, the pooled Pearson correlation between the estimated and actual femoral neck BMD was \textbf{$r=0.760$} (95\% CI 0.695--0.812; Fisher's $z$; 10 folds, $N=233$).

\begin{figure*}
    \centering
    \includegraphics[width=1\linewidth]{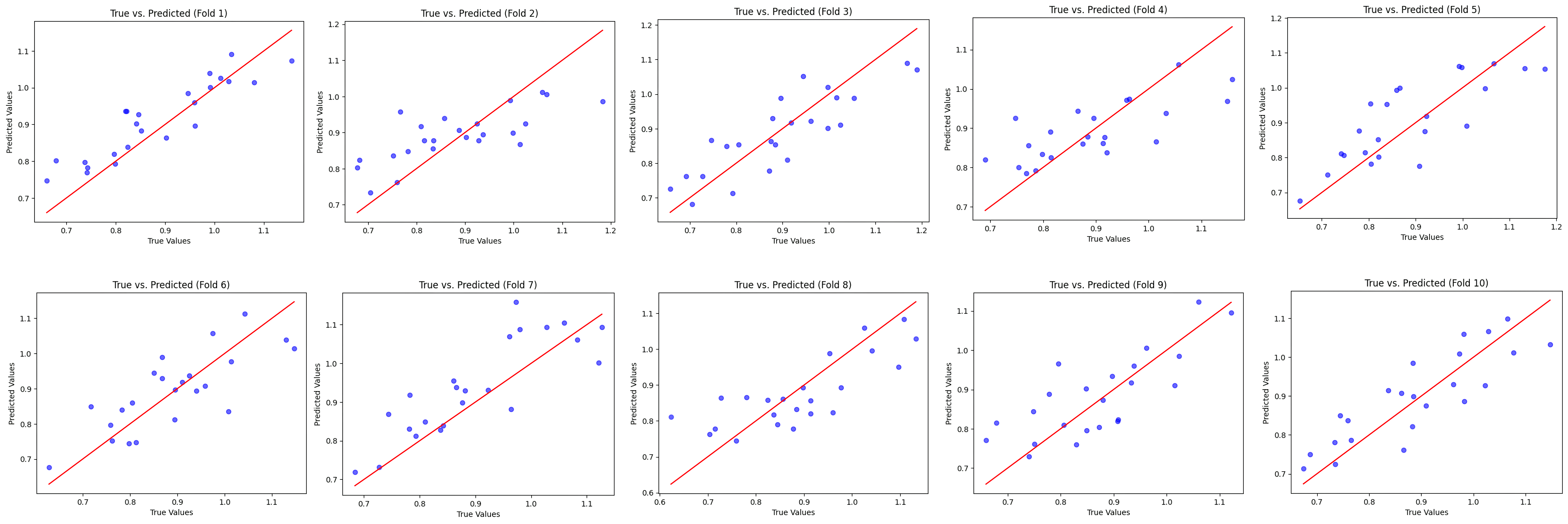}
    \caption{True vs. estimated femoral neck BMD for each fold in 10-fold cross-validation. Each subplot shows the estimated values ($\hat{y}$) against the reference DXA values ($y$), the red line is the identity line ($y=x$).}
    \label{fig:image-True-vs-predicted-femoral-neck-BMD}
\end{figure*}

In addition to the average performance, the model demonstrated consistent behavior across all folds. As shown in Figure~\ref{fig:image-True-vs-predicted-femoral-neck-BMD}, the $R^2$ values for most folds were around 0.70. Notably, fold 1 achieved the best performance (MSE: 0.0032, MAE: 0.0497, $R^2$: 0.8434), followed by fold 10 (MSE: 0.0040, MAE: 0.0595, $R^2$: 0.8029) and fold 3 (MSE: 0.0049, MAE: 0.0642, $R^2$: 0.7919). Even in the lowest-performing fold (fold 2, MSE: 0.0077, MAE: 0.0651, $R^2 = 0.5899$), the model preserved a clear linear trend, indicating stable performance under more challenging data splits.These results demonstrate consistent regression across different dataset partitions. 

\subsubsection{T-test Results}
To evaluate the statistical significance of the performance improvement, we performed paired t-tests on the proposed cross-attention model and the baseline methods (image only, metadata only, and direct fusion) to evaluate the MSE, MAE, and $R^2$ metrics on 10-fold cross-validation of each experiment. The results are summarized in Table~\ref{tab:t-test_results}.

For MSE and MAE, negative t-values indicate that the error of the cross-attention model is significantly lower than the baseline method. For the $R^2$ metric, positive t-values indicate that the explained variance of the cross-attention model is significantly higher than the baseline method. All observed $p$ values are below 0.05, confirming that the performance difference is statistically significant, revealing the effectiveness of utilizing a cross-attention mechanism to fuse features from different modalities.

\begin{table*}[htbp]
\footnotesize
\centering
\caption{Statistical significance (t-test) results comparing cross-attention model with baselines, including Image-only model, metadata-only}
\label{tab:t-test_results}
\begin{tabular}{lccc}
\hline
\textbf{Comparison} & \textbf{MSE (t, p)} & \textbf{MAE (t, p)} & \textbf{$R^2$ (t, p)} \\
\hline
Cross-attention vs Image-Only   & ($-3.63$, $0.0055$) & ($-3.71$, $0.0048$) & ($5.15$, $0.0006$) \\
Cross-attention vs Meta-Only    & ($-5.79$, $0.0003$) & ($-11.61$, $0.0000$) & ($6.52$, $0.0001$) \\
Cross-attention vs Direct Fusion& ($-2.70$, $0.0243$) & ($-2.28$, $0.0487$) & ($4.69$, $0.0011$) \\
\hline
\end{tabular}
\end{table*}

\subsubsection{Field-level Attention Visualization}
\label{sec:attention_viz}
\begin{figure*}
    \centering
    \includegraphics[width=1\linewidth]{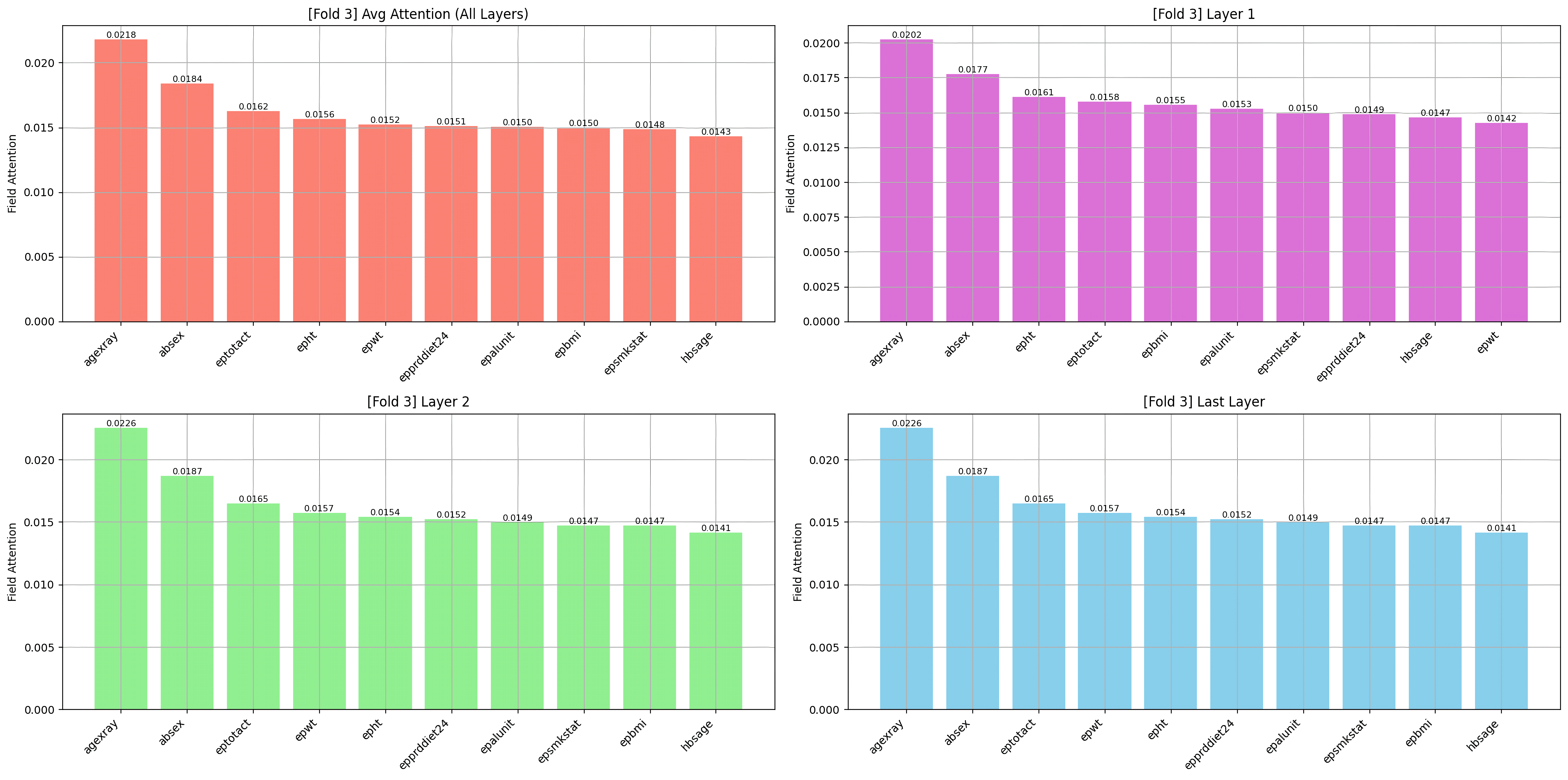}
    \caption{Field-level attention visualization for Fold~3 across all layers, with "agexray" and "absex" receiving the highest attention on average. The definitions and descriptions of the metadata variables on the x-axis are provided in Table~\ref{tab:metadata_summary}.}
    \label{fig:image-field_attention_fold3}
\end{figure*}

As the attention in our model operates on the vector representations of image and metadata rather than spatial image tokens, it is not feasible to visualize image attention maps. Therefore, we focus our analysis on the field-level attention assigned to metadata fields across layers. To explore how the proposed model gradually refines its use of metadata fields through stacked cross-attention layers, the field-level attention weights of Fold 3 at different layers are visualized as shown in Figure \ref{fig:image-field_attention_fold3}. Each plot corresponds to the average attention of a particular layer over all heads. The top left shows the overall average of all layers, and the remaining subplots show the attention distribution of layer 1, layer 2, and the third (last) layer, respectively.

We observe that the attention weights gradually become more selective with the number of layers. In layer 1, the attention is more evenly distributed across fields, indicating that the model is still in the early stages of incorporating metadata. In layer 2, biologically relevant fields (e.g., age at taking X-ray and sex) receive more attention, while fields with less contribution start to receive less attention. In the last layer, the trend has been identified, with popular fields (age at taking x-ray, sex, and physical activity in last 2 weeks) dominating the attention, indicating a refinement of the feature selection process, with more attention paid to high-impact metadata. Notably, age, sex and physical activity are well-known clinical risk factors for low BMD \cite{lane2006epidemiology}, therefore, their dominance in the later layers suggests that the model has learned to prioritize clinically meaningful variables. This alignment with established medical knowledge enhances the interpretability and credibility of the attention mechanism.

This incremental improvement suggests that the model performs hierarchical feature filtering, learning which metadata dimensions are most predictive through successive attention updates, thus enhancing its interpretability.
\subsection{Perturbation Test}
\label{sec:perturbation}
To further evaluate the reliability of the proposed model to image variability, we performed perturbation tests on the three best-performing folds (fold 1, fold 3, and fold 10) in the 10-fold cross-validation. In the cross-validation stage, only the original test images were used for performance evaluation. In this experiment, we introduced additional perturbations to the test images to simulate the image changes that may occur in the real world and verify the regression stability of the model under these conditions.

\begin{figure*}
    \centering
    \includegraphics[width=1\linewidth]{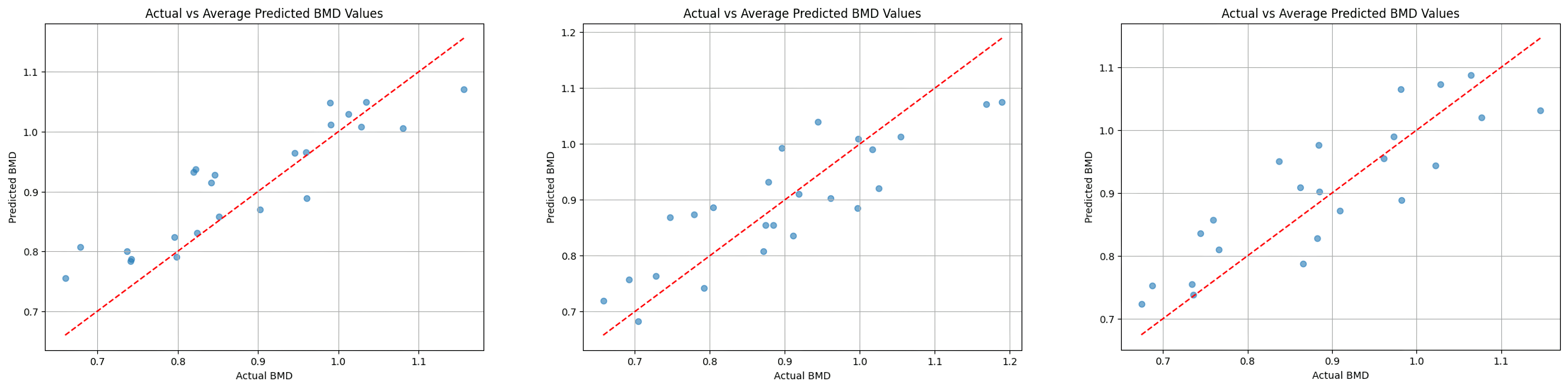}
    \caption{Scatter plots of actual vs. average \emph{estimated} BMD under perturbation testing (left to right: folds 1, 3, and 10). Each point is the mean over 20 augmented variants of a single test image; the red line is the identity line ($y=x$).}
\label{fig:perturbation_scatter}
    \label{fig:image-Perturbation-test}
\end{figure*}

Specifically, for the test samples in each of the above folds, we used the similar image augmentation strategy as the training stage (see Table~\ref{tab:training-augmentations} for details), including random brightness/contrast adjustment, affine transformation, Gaussian noise, etc., to generate 20  augmented variants of each image, where 1 to 2 augmentation methods were randomly selected and applied for each perturbation. These perturbed images were then input into the model checkpoints obtained from the corresponding fold training for estimation and evaluated using the same regression metrics as the main experiment, including mean squared error (MSE), mean absolute error (MAE), and determination coefficient ($R^2$).
Figure~\ref{fig:image-Perturbation-test} and Table~\ref{tab:perturbation-test} together show the results of perturbation testing on the best-performing folds (1, 3, and 10). The scatter plot in Figure~\ref{fig:perturbation_scatter} shows the relationship between the actual BMD values and the average predicted values for 20 augmented versions of each test image. Despite the large number of perturbations, the predicted values remain consistent with the identity line, indicating that the model retains its stability and can capture the underlying BMD pattern.

As shown in Table~\ref{tab:perturbation-test}, the model performance only slightly degrades under perturbations. The average $R^2$ score for the three folds drops slightly compared to the original value, while the MAE and MSE remain relatively low. This confirms that the model is resilient to variations introduced by different perturbations, such as brightness shifts, geometric distortions, and noise.
\begin{table*}[htbp]
\centering
\caption{Performance comparison on original vs. perturbed test sets for the top 3 folds}
\label{tab:perturbation-test}
\begin{tabular}{lcccccc}
\hline
\textbf{Fold} & \multicolumn{3}{c}{\textbf{Original Test Set}} & \multicolumn{3}{c}{\textbf{Perturbation Test}} \\
\cmidrule(lr){2-4} \cmidrule(lr){5-7}
& MSE & MAE & $R^2$ & MSE & MAE & $R^2$ \\
\hline
Fold 1  & 0.0032 & 0.0497 & 0.8434 & 0.0039 & 0.0492 & 0.7645 \\
Fold 3  & 0.0049 & 0.0642 & 0.7919 & 0.0053 & 0.0641 & 0.7268 \\
Fold 10 & 0.0040 & 0.0595 & 0.8029 & 0.0045 & 0.0578 & 0.7376 \\
\hline
\end{tabular}
\end{table*}

\subsection{Clinical Screening Performance}
To assess the screening potential of the regression model, we converted the continuous predicted BMD values into binary labels using a clinically relevant threshold. T-score is a standardized score that quantifies how much an individual’s BMD deviates from the mean value of a young, healthy reference population, and is strongly correlated with femoral neck BMD. In this study, the T-score for the femoral neck was calculated as:

\begin{equation}
\text{T-score} = \frac{\text{BMD}_{\text{patient}} - \text{BMD}_{\text{Mean of Young Adult Reference}}}{\text{SD}_{\text{Young Adult reference}}}
\end{equation}
where $\mathrm{BMD}_{\text{Mean of Young Adult Reference}}$ and $\mathrm{SD}_{\text{Young Adult reference}}$ refer to the mean and standard deviation of femoral neck BMD from a healthy young adult reference population. The formula for T-score calculation and the reference values ($\mathrm{BMD}_{\text{Mean of Young Adult Reference}} = 1.038~\mathrm{g/cm}^2$ and $\mathrm{SD}_{\text{Young Adult reference}} = 0.139~\mathrm{g/cm}^2$), which used NHANES III data as reference with the appropriate adjustment for the DXA scanner used. And $\mathrm{BMD}_{\text{patient}}$ refers to the BMD value of the femoral neck of the individual under evaluation.

Due to the limited sample size, our dataset contains very few subjects in the T-score $< -2.5$ (femoral neck BMD $< 0.691~\mathrm{g/cm}^2$), we adopted T-score $< -1$ (femoral neck BMD $< 0.899~\mathrm{g/cm}^2$) as the threshold for classifying low bone mass. Accordingly, subjects were dichotomized into low and normal BMD groups according to this threshold. In total, there are 131 patients with femoral neck BMD  $< 0.899$, around 56.22\% of the whole dataset. This binary classification enabled evaluation of the model’s potential for clinical screening of low BMD. 

\begin{figure*}
    \centering
    \includegraphics[width=1\linewidth]{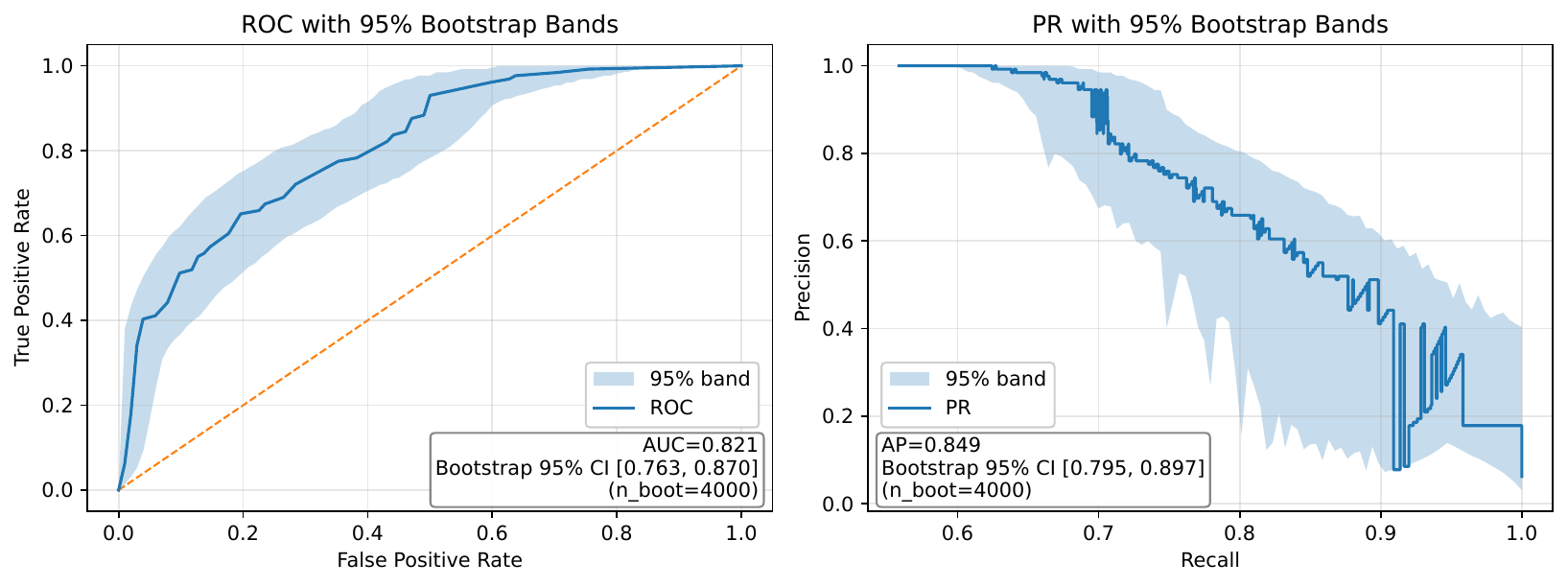}
    \caption{ROC and PR curves for binary classification (clinical threshold: femoral neck BMD $< 0.899$). Shaded areas indicate pointwise 95\% stratified-bootstrap bands ($n\_{\text{boot}}=4000$).}
    \label{fig:roc_pr_curves}
\end{figure*}

The classification performance was assessed using ROC and precision-recall curves in Figure~\ref{fig:roc_pr_curves}, with the area under the ROC curve (AUC) and average precision (AP) used as summary statistics. These metrics demonstrate the ability of the model to discriminate individuals with low bone mass, which is of direct clinical relevance for early osteoporosis screening.

\begin{figure*}
    \centering
    \includegraphics[width=1\linewidth]{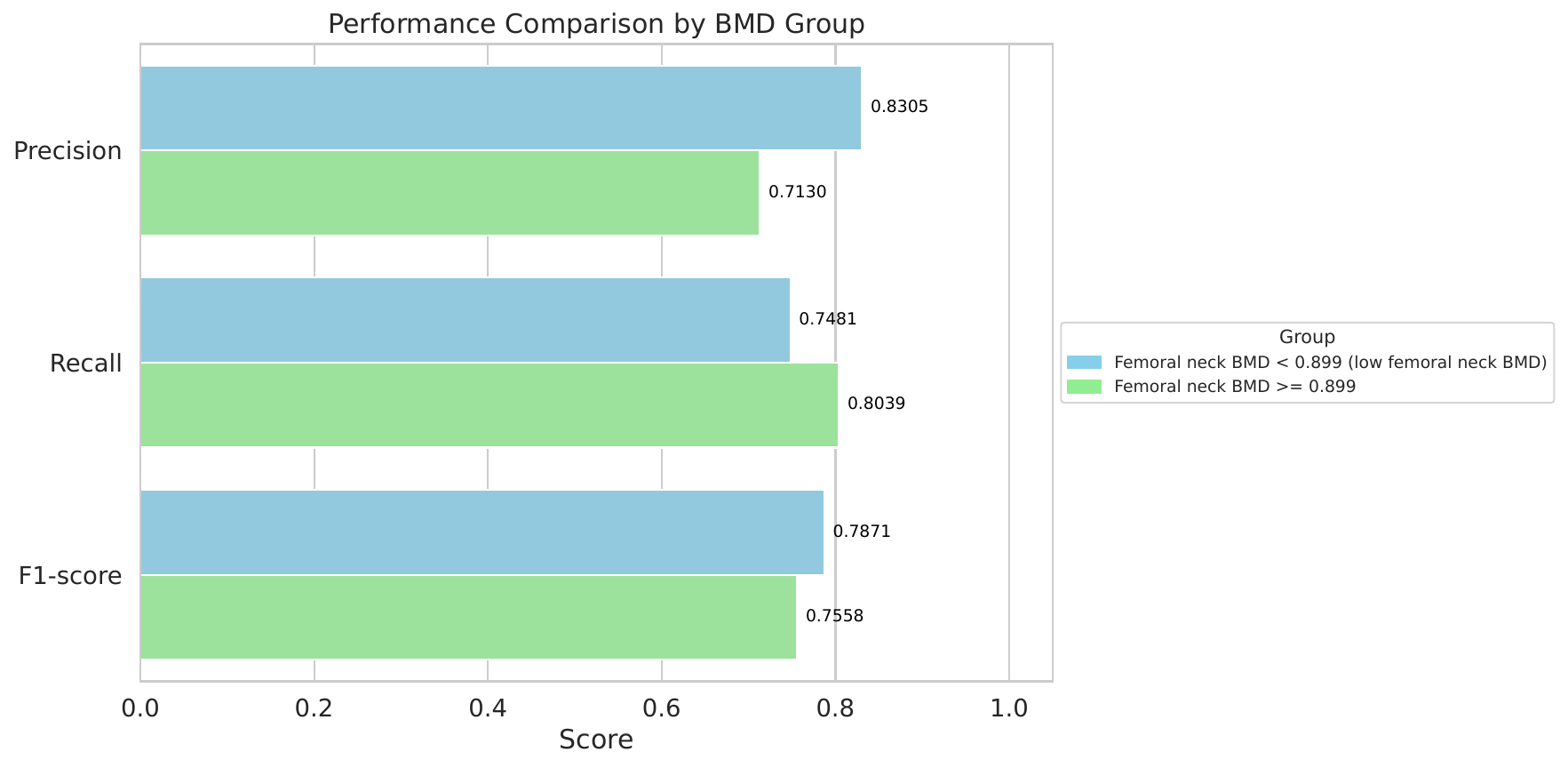}
    \caption{Bar plot of precision, recall, and F1-score for clinical binary classification of
low femoral neck BMD (clinical threshold: femoral neck BMD $< 0.899)$.}
    \label{fig:bar-plot}
\end{figure*}

Figure~\ref{fig:bar-plot} illustrates the precision, recall, and F1-score for binary classification based on the clinical threshold of femoral neck BMD $<0.899~\mathrm{g/cm}^2$. For the low BMD group, the model achieved a precision of 0.8305, a recall of 0.7481, and an F1-score of 0.7871, indicating that this model has the potential to identify most predicted cases of low BMD. For the normal BMD group, the model achieved precision of 0.7103, recall of 0.8039 (which equals specificity, with 95\% CI by Wilson: 0.716–0.869), and F1-score of 0.7558, still suggesting credible performance in identifying individuals without low bone mass.

In summary, across all subjects, the model achieved an overall accuracy of 0.7725, weighted-avg precision of 0.8305, recall of 0.7481, and F1-score of 0.7871, indicating reliable discrimination between subjects with and without low bone mass.

\subsection{Comparative Evaluation on Our Cohort}
\label{Comparative evaluation on our cohort}
Direct head-to-head comparison with previous studies is to some extent infeasible because most clinical datasets are not publicly available and protocols differ across works. To ensure a fair assessment, we re-implemented representative prior models (see Table~\ref{tab:feature_matrix}) and trained/evaluated them on our cohort under identical conditions: the same input (hip X-ray ROI + metadata), the same preprocessing/augmentations, the same optimization (Adam; Weighted Smooth L1 loss), and stratified 10-fold cross-validation with a fixed random seed (42). For the screening task, the positive class is defined by the clinical threshold of femoral neck $\mathrm{BMD}<0.899~\mathrm{g/cm^2}$. Results are reported as mean~$\pm$~SD across folds. All baselines were implemented following the descriptions in the cited studies and trained/evaluated on this cohort. Where a cited study lacked specific implementation details, we resolved ambiguities using widely adopted defaults or the closest stated settings, while keeping identical data splits and training budgets across methods. The values reported in Table~\ref{tab:comparative_on_our_data} are from our experiments. XAttn-BMD consistently outperforms the representative baselines in both screening and BMD estimation, indicating that cross-attention adaptively integrates imaging and metadata in an anatomically and clinically meaningful way.

\paragraph*{BMD estimation comparison (regression)}
Across late-concatenation CNN baselines (InceptionV3, ResNet18/50, VGG16, and a VGG16+self-attention variant conceptually similar to prior work), XAttn-BMD attains the lowest MSE/MAE and the highest $R^2$, consistently outperforming alternatives under the same training budget.

\paragraph*{Screening performance comparison(binary classification)}
Our cross-attention model XAttn-BMD achieves higher accuracy and F1-score than a strong CNN baseline (DenseNet121), indicating a better balance between sensitivity and precision on this imbalanced task.

\begin{table*}[t]
\centering
\caption{Comparative performance on our cohort (stratified 10-fold cross-validation; mean~$\pm$~SD). }
\label{tab:comparative_on_our_data}

\begin{subtable}{0.50\textwidth}
  \centering
  \caption{\footnotesize Screening performance comparison}
  \begin{adjustbox}{max width=\linewidth}
    \begin{threeparttable}
      \setlength{\tabcolsep}{4pt}\renewcommand{\arraystretch}{0.95}
      \begin{tabular}{@{}lcc@{}}
        \toprule
        \textbf{Model} & \textbf{Accuracy} & \textbf{F1-score}\\
        \midrule
        \makecell[l]{\textbf{XAttn-BMD} (ours)}  & \textbf{0.77} & \textbf{0.79}\\
        \addlinespace[6pt]
        \makecell[l]{Feng (2023)\cite{feng2023deep}\textsuperscript{1}}
          & 0.74 & 0.71\\
        \bottomrule
      \end{tabular}
      \vspace{-2pt}
      \begin{tablenotes}[flushleft]
        \footnotesize
        \item[$^1$] Compared with VGG16 and ResNet50, DenseNet121 was reported by \cite{feng2023deep} as the best-performing classification model on their dataset, we therefore use DenseNet121 here for comparison.
      \end{tablenotes}
    \end{threeparttable}
  \end{adjustbox}
\end{subtable}

\vspace{0.8em}

\begin{subtable}{\textwidth}
  \centering
  \caption{\footnotesize BMD estimation comparison}
  \begin{adjustbox}{max width=\textwidth, center}
    \begin{threeparttable}
      \begin{tabular}{lccc}
        \toprule
        \multicolumn{1}{l}{\textbf{Model}} & \textbf{MSE} & \textbf{MAE} & $\mathbf{R^2}$ \\
        \midrule
        \makecell[l]{\textbf{XAttn-BMD} (ours) } 
          & \textbf{0.0055 $\pm$ 0.0012} 
          & \textbf{0.0592 $\pm$ 0.0059} 
          & \textbf{0.7011 $\pm$ 0.0802} \\
        \addlinespace[6pt]
        \makecell[l]{Golestan (2023)\cite{golestan2023approximating}}
          & 0.0078 $\pm$ 0.0017 & 0.0703 $\pm$ 0.0069 & 0.5327 $\pm$ 0.0870 \\
        \addlinespace[6pt]
        \makecell[l]{Ho (2021)\cite{ho2021application}}
          & 0.0071 $\pm$ 0.0022 & 0.0670 $\pm$ 0.0101 & 0.5759 $\pm$ 0.1136 \\
        \addlinespace[6pt]
        \makecell[l]{Hsieh (2021)\cite{hsieh2021automated}\textsuperscript{2}}
          & 0.0141 $\pm$ 0.0040 & 0.0958 $\pm$ 0.0160 & 0.1409 $\pm$ 0.2685 \\
        \addlinespace[6pt]
        \makecell[l]{Sato (2022)\cite{sato2022deep}}
          & 0.0071 $\pm$ 0.0017 & 0.0672 $\pm$ 0.0094 & 0.5710 $\pm$ 0.0897 \\
        \addlinespace[6pt]
        \makecell[l]{Wang (2022)\cite{wang2022lumbar}\textsuperscript{3}}
           & 0.0083 $\pm$ 0.0020 & 0.0722 $\pm$ 0.0091 & 0.5000 $\pm$ 0.1102 \\
        \bottomrule
      \end{tabular}

      \begin{tablenotes}[flushleft, para]
        \footnotesize
        \item[$^2$] Since \cite{hsieh2021automated} trained VGG16 on hip X-ray images (see Table~\ref{tab:feature_matrix}), we use VGG16 here for comparison rather than ResNet34.
        
        \item[$^3$] We approximate the attention mechanism from \cite{wang2022lumbar} using grid-based pseudo-ROIs with self-attention, as manual anatomical annotation for ROI detection was not feasible with our cohort size. This methodological difference may limit the direct comparability with their original approach.
      \end{tablenotes}
    \end{threeparttable}
  \end{adjustbox}
\end{subtable}

\end{table*}

\section{Discussion}
Our findings show that XAttn-BMD delivers robust, interpretable femoral neck BMD estimation in a moderately sized single-center cohort. This addresses a persistent gap in prior work, which largely relied on image-only models trained on large, multicenter datasets or included only a few metadata variables, with limited attention to explaining clinical factors. However, due to differences in model design, datasets, imaged anatomical sites, clinical endpoints, and data access across existing approaches (see Table~\ref{tab:feature_matrix}), direct and fair comparisons remain challenging. This structured qualitative summary analyzes the characteristics of each method and emphasizes the importance of developing a scalable approach for estimating femoral neck BMD, particularly in resource-constrained settings where interpretable estimates are required.

Technically, our model performs dynamic, bidirectional interaction between images and metadata features, leveraging their complementary strengths. Field-level attention weights provide factor-specific interpretability, quantifying each clinical variable’s contribution and improving transparency in clinical decision-making. A clinically tailored loss function further aligns optimization with diagnostic needs and improves performance under class imbalance.

Clinically, our approach demonstrates practical value in resource-limited settings where large-scale data collection is infeasible. The model's attention to established risk factors (age, sex, physical activity state, etc.) aligns with clinical knowledge, while novel attention patterns may reveal underexplored biomarkers worthy of further investigation.

Despite these strengths, several limitations remain. First, the dataset is relatively small and single-center, which may limit generalizability. Second, the BMD distribution is imbalanced, with relatively few extreme cases, potentially affecting performance in the tails. Third, although we provide clinical factor-level interpretability, both performance and explanations warrant external validation on independent, multicenter cohorts. Future work will pursue such validation and integration with electronic health records to support real-world deployment.

\section{Conclusion}
This study presents XAttn-BMD, a multimodal framework based on a cross-attention mechanism for estimating femoral neck bone mineral density from hip X-ray images and health-related metadata. The cross-attention module enables dynamic interactions between image and metadata features, enhancing their representations. In addition, a Weighted Smooth L1 loss is designed to emphasize clinically significant BMD ranges.

Experimental results on a real-world cohort demonstrate that the proposed method outperforms naive feature concatenation without cross-attention, achieving superior performance across multiple metrics. Furthermore, our results highlight the significance of integrating metadata, which provides complementary clinical context, enables factor-level interpretability, and notably improves predictive performance. These findings highlight the effectiveness of cross-attention-based multimodal fusion and a tailored loss function for clinical BMD estimation, even on a moderately sized dataset. In addition, when applied to a clinically relevant binary classification setting, the proposed model also achieved considerable screening performance, further demonstrating its potential utility for early identification of individuals at risk of low bone mass and osteoporosis.

Nevertheless, it should be noted that the present study is limited by the relatively small sample size and the use of data from a single cohort, which may affect the generalizability of the findings. Future work will focus on external validation using larger, multicenter datasets, integrating additional imaging modalities (e.g., DXA, CT), and incorporating explainable AI techniques to enhance interpretability and clinical trust. In addition, although our main task is to estimate bone mineral density, estimated values have the potential to be used for risk stratification in clinical practice. Therefore, future studies may also focus on systematically evaluating the model's ability to identify high-risk individuals using established bone density thresholds.

\section*{CRediT authorship contribution statement}
\textbf{Yilin Zhang}: Conceptualization; Methodology; Software; Validation; Formal analysis; Investigation; Data curation; Visualization; Writing – Original Draft; Writing – Review \& Editing.
\textbf{Leo D. Westbury}: Data curation; Writing – Review \& Editing.
\textbf{Elaine M. Dennison}: Data curation; Writing – Review \& Editing.
\textbf{Nicholas C. Harvey}: Data curation; Writing – Review \& Editing.
\textbf{Nicholas Fuggle}: Supervision; Conceptualization, Writing – Review \& Editing.
\textbf{Rahman Attar}\textsuperscript{\corref{cor1}}: Supervision; Conceptualization; Methodology; Project administration;  Resources; Writing – Review \& Editing.

\section*{Ethical background}
Ethical approval for this study was obtained from the Health and Social Care Research Ethics Committee B (HSC REC B), Health Research Authority (REC reference: 25/NI/0124; IRAS project ID: 359237). All data were anonymized before analysis, and informed consent was obtained at the time of data collection. The baseline Hertfordshire Cohort Study had ethical approval from the Hertfordshire and Bedfordshire Local Research Ethics Committee and the follow-up had ethical approval from the East and North Hertfordshire Ethical Committees. 

\section*{Declaration of competing interest}
The authors declare that they have no known competing financial interests or personal relationships that could have appeared to influence the work reported in this paper.

\section*{Data availability}
The hip radiographs and accompanying clinical metadata analysed in this study were obtained from the Hertfordshire Cohort Study (HCS). These are third-party data collected under ethics approvals and participant consent that do not permit open public sharing of individual-level medical images or metadata. 

\section*{Funding sources}
RA and NRF wish to acknowledge funding from the Medical Research Council (MRC) and National Institute for Health and Care Research (NIHR).

\section*{Acknowledgments}
The authors would like to thank the clinical collaborators for their valuable assistance with data preparation and interpretation. We also appreciate the constructive feedback and suggestions provided by colleagues during the course of this work.

\bibliographystyle{elsarticle-num}
\bibliography{references}

\end{document}